\documentclass[lettersize,journal]{IEEEtran}
\usepackage{amsmath,amsfonts}
\usepackage{algorithmic}
\usepackage{algorithm}
\usepackage{array}
\usepackage[caption=false,font=normalsize,labelfont=sf,textfont=sf]{subfig}
\usepackage{textcomp}
\usepackage{stfloats}
\usepackage{url}
\usepackage{verbatim}
\usepackage{graphicx}
\usepackage{cite}
\usepackage{booktabs}
\usepackage{orcidlink}
\usepackage{tabularx}
\usepackage{multirow}

\begin{document}

\title{Efficient and Scalable Point Cloud Generation with Sparse Point-Voxel Diffusion Models}
%Efficient and Scaleable Point Cloud Generative Modeling through Sparse Point-Voxel Diffusion Models} %Tasks}
%\title{Sparse Point-Voxel Diffusion Models for Efficient and Scaleable Point Cloud Generative Modeling}
%\title{Sparse Point-Voxel Convolutions for Efficient And Scaleable Point Cloud Diffusion Models}
%\title{Efficient and Versatile Point Cloud Diffusion via Sparse Point-Voxel Models}

% \author{Ioannis Romanelis}
% \author{Vlassis Fotis}
% \author{Athanasios Kalogeras}
% \author{Konstantinos Moustakas,~\IEEEmembership{Staff,~IEEE,}
% \author{Adrian Munteanu}
\author{Ioannis Romanelis \orcidlink{0000-0002-2917-8705}, Vlassios Fotis \orcidlink{0000-0002-1212-5500}, Athanasios Kalogeras \orcidlink{0000-0001-5914-7523},~\IEEEmembership{Senior Member,~IEEE}, Christos Alexakos \orcidlink{0000-0002-8932-6781},~\IEEEmembership{Member,~IEEE} , Konstantinos Moustakas \orcidlink{0000-0001-7617-227X},~\IEEEmembership{Senior Member,~IEEE}, Adrian Munteanu \orcidlink{0000-0001-7290-0428},~\IEEEmembership{Member,~IEEE} 
        % <-this % stops a space
%\thanks{This paper was produced by the IEEE Publication Technology Group. They are in Piscataway, NJ.}% <-this % stops a space

\thanks{Ioannis Romanelis, Vlassis Fotis, and Konstantinos Moustakas are with the Department of Electrical and Computer Engineering, University of Patras. Ioannis Romanelis, Vlassis Fotis, Athanasios Kalogeras, and Christos Alexakos are with the Industrial Systems Institute (ISI) - Athena Research Center. Ioannis Romanelis and Adrian Munteanu are with the Department of Electronics and Informatics, Vrije Universiteit Brussel}
%\thanks{Manuscript received April 19, 2021; revised August 16, 2021.}
}



\maketitle

\begin{abstract}

We propose a novel point cloud U-Net diffusion architecture for 3D generative modeling capable of generating high-quality and diverse 3D shapes while maintaining fast generation times.
Our network employs a dual-branch architecture, combining the high-resolution representations of points with the computational efficiency of sparse voxels. Our fastest variant outperforms all non-diffusion generative approaches on unconditional shape generation, the most popular benchmark for evaluating point cloud generative models, while our largest model achieves state-of-the-art results among diffusion methods, with a runtime approximately $\textbf{70\%}$ of the previously state-of-the-art PVD.
Beyond unconditional generation, we perform extensive evaluations, including conditional generation on all categories of ShapeNet, demonstrating the scalability of our model to larger datasets, and implicit generation which allows our network to produce high quality point clouds on fewer timesteps, further decreasing the generation time. %allowing for a friendly user experience is real applications. 
Finally, we evaluate the architecture's performance in point cloud completion and super-resolution. Our model excels in all tasks, establishing it as a state-of-the-art diffusion U-Net for point cloud generative modeling. The code is publicly available at 
 \url{https://github.com/JohnRomanelis/SPVD.git}.
\end{abstract}


\section{Introduction}
Generative models have emerged as powerful tools in the realm of artificial intelligence, offering significant advances in the automated generation of digital content across various modalities, such as text \cite{brown2020language}, music \cite{9132664, copet2024simple, lam2024efficient}, image \cite{DDPM, DDIM, DiT} and video\cite{ho2022video}. Recently, the focus has extended to 3D generative models which hold promise for a wide array of applications like computer vision, computer graphics, and robotics. 
Among 3D data representations, point clouds are becoming increasingly common. %one of the most common representations of 3D data.
This is primarily because they are the direct output of 3D sensors, increasing the availability of data, and can express a higher level of detail compared to other representations such as voxel grids. Additionally, a plethora of algorithms have been developed to transform point clouds into more sophisticated representations, such as 3D meshes \cite{Point2Mesh, siren, hashimoto2019normal, 9406392, Shape-as-points, chen2024meshanything}.
%The main reasons include that they are the direct output of 3D sensors, they can express a higher level of detail than the voxel grids and several algorithms have been implemented to transform point clouds to more sophisticated representations such as 3D meshes. 

\begin{figure}
    \centering
    \includegraphics[width=0.99\linewidth]{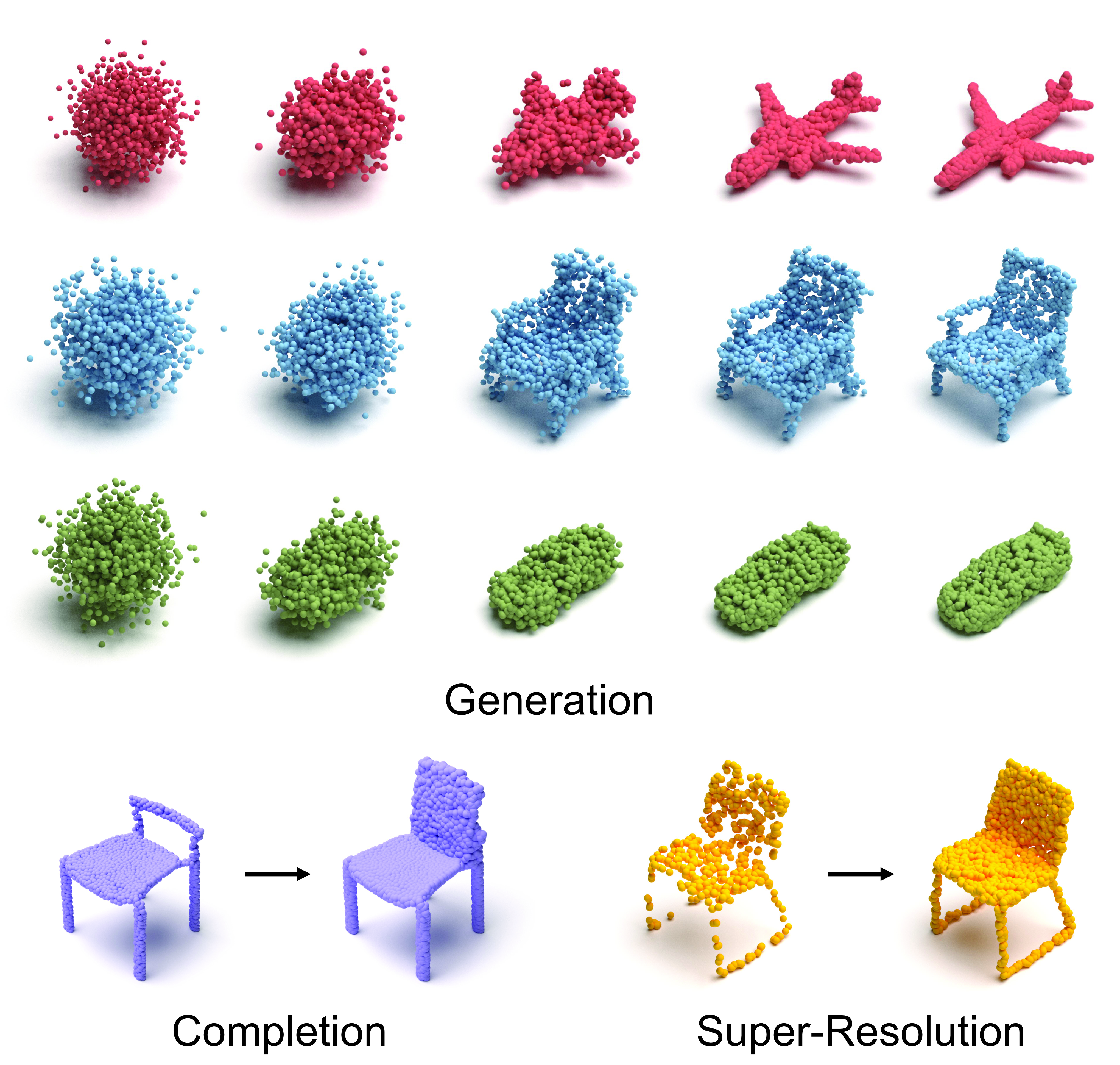}
    \caption{The proposed Sparse Point-Voxel Diffusion (SPVD) is a novel diffusion architecture designed for efficient and scalable point cloud generation tasks. The generation process visualizes the gradual transformation of a noisy sample into a clean 3D shape. The completion and super-resolution tasks further demonstrate the capabilities of the proposed architecture.}
    \label{fig:graphicalAbstract}
\end{figure}

Several methods have been proposed for point cloud generation, employing various architectures and pipelines, such as VAEs \cite{SetVAE, FoldingNet}, GANs \cite{achlioptas18a, valsesia2018learning, 9009495}, Gradient Fields \cite{ShapeGF}, Normalizing Flows 
\cite{PointFlow, softflow, klokov20eccv}, and Diffusion models \cite{PVD, DPM, LION}.
Diffusion models \cite{DDPM}, recognized as the State-of-the-Art generative approach in visual computing (including image, video), are gaining increasing popularity in the Point Cloud domain due to their ability to produce high-fidelity and diverse shapes. These models operate by progressively denoising a noise sample from a Gaussian distribution to generate a clean novel shape. % sampling a noise sample from a Normal distribution and progressively denoising it to generate a novel shape. 
%However, this process can be quite time-consuming, as it requires hundreds of network activations. 
However, thousands of steps are required for this generation process \cite{PSF}, equaling to thousands of network activations. 

%Moreover, Point Cloud models are inherently slow in opertion, typically requiring multiple point sampling and grouping operations, to perform point convolutions. Due to the lack of straighforward methods for sampling point neighbors, slower k-nearest neighbors (k-NN) methods should be applied. Furthermore, most point cloud models lack the ability to scale to larger inputs without significant design changes in their architecture. Therefore, designing an architecture that can handle varying numbers of input points is also challenging.

%\IEEEpubidadjcol

Moreover, current state-of-the-art point-based Point Cloud models are inherently slow, with up to 90\% of their runtime dedicated to structuring irregular data rather than actual feature extraction \cite{pvcnn}. This structuring includes sampling and neighbor searching operations necessary for tasks like downsampling the point cloud, forming point neighborhoods, and interpolating point features. Unlike image pixels, which are organized in a fixed grid making these operations straightforward, point clouds require more complex algorithms such as furthest point sampling and k-nearest neighbor searches in continuous space. While voxel-based methods could potentially overcome these speed limitations, they suffer from significant information loss due to the aggressive downsampling needed to manage the cubical memory demands of voxel grids \cite{spvnas}. Additionally, voxel-based methods have been shown to produce poor generation results \cite{PVD}.

We propose SPVD, a novel UNet diffusion architecture that achieves state-of-the-art generation results while significantly reducing generation runtime compared to other diffusion methods. Our model, inspired by \cite{spvnas}, combines a sparse voxel backbone designed to efficiently extract neighboring information with a high-fidelity point branch that preserves the fine details of the points. To minimize hard-to-interpret design choices, our sparse voxel backbone follows the DDPM \cite{DDPM} UNet architecture, incorporating only domain-specific adaptations. Additionally, to further decrease the generation time, we modify the voxelization pipeline from \cite{spvnas} to run entirely on the GPU. To put the numbers into perspective, PVD \cite{PVD}, the current state-of-the-art, requires more than 1 hour to generate 662 samples, which is the size of the ShapeNet - Chair category test set. In contrast, our model's fastest variant completes this task in less than 15 minutes, while the largest variant does not exceed 45 minutes\footnote{All time measurements were performed on the same machine using a NVIDIA RTX 3090 GPU.}. Moreover, all versions of our network have been trained on a 24GB VRAM GPU, making it accessible for retraining and use by the academic community without the need for expensive and often unavailable equipment.

We quantitatively evaluate our model, following the paradigm of previous works \cite{achlioptas18a, softflow, klokov20eccv, DPM, ShapeGF, PointFlow, PSF, PVD}, by measuring the 1-NN metric for the unconditional generation results in the Car, Airplane and Chair categories of ShapeNet. To test the model's scalability with increasing amounts of data, we train a variant for conditional generation on all categories of ShapeNet. To further decrease the generation time we study the DDIM generation rule \cite{DDIM}, which allows generation with fewer iteration, and its effect on the generation quality. Finally, we qualitatively test our model in other candidate tasks, such as completion and super-resolution. 
%In addition to generation, we evaluate our model on shape completion and point cloud super resolution. We implement both a conditional and an unconditional variant of the model. Furthermore, we employ the DDIM generation scheduler to further reduce generation runtime by decreasing the required number of denoising iterations. Finally, we perform several ablation studies to examine the performance of different network sizes, various methods of incorporating the point branch, and how these variations facilitate the training process.

To summarize our contributions are the following: 
\begin{itemize}
    \item We propose SPVD, a novel diffusion U-Net  architecture that combines the point representations with sparse voxels for efficient point cloud processing. 
    \item We achieve state-of-the-art results in the most common generative benchmark - unconditional generation on Airplane, Chair, Car categories of ShapeNet - while reducing generation time compared to the previous state-of-the-art diffusion model.
    \item We present extensive quantitative and qualitative results to demonstrate our model's ability to scale to larger datasets, generate shapes faster through implicit generation, and perform additional generative tasks such as shape completion and super-resolution.
    
\end{itemize}

\section{Related Work}

\subsection{Deep Learning for Point Clouds}
Designing an effective 3D generative model requires selecting a backbone architecture that balances execution speed with accuracy. PointNet \cite{pointnet} utilizes shared-MLPs across points to project them into higher dimensions and extract global features through pooling; however, it lacks on descriptive power due to the absence of neighbor feature aggregation. PointNet++ \cite{pointnet++} addresses this by applying PointNets withing small point neighborhoods to extract local features. Subsequent studies \cite{rsconv, kpconv, dgcnn, spidercnn, PointWavelet, point_transformer}  explore different kernels for neighborhood feature aggregation. While these methods achieve state-of-the-art results in classification and segmentation tasks, their execution time is constrained by the time required for point sampling and grouping operations.

Transformer-based models \cite{PointBERT, PointMAE, PointM2AE, ExpPointMAE, PointGPT, Point2Vec} create point patches and process them using transformer blocks \cite{AttentionIsAllYouNeed}. Although these models reduce the number of point operations, computing the attention matrices remains a time-intensive process. 

%Point-Voxel CNNs \cite{pvcnn} introduce a hybrid convolution approach, combining a high resolution point branch with a low-resolution voxel branch to aggregate neighborhood information. The authors propose two network architectures in their manuscript. The first one, PVCNN, uses a series of Point Voxel CNNs to extract point features. The second variant, PVCNN++, inspired by PointNet++, also utilizes point operations. This means that after each voxel convolution, the points are sampled and grouded into neightborhoods, and collective features are extracted for the neightborhood centroids, thus creating a point encoder with decreasing number of points. While this approach increases the networks ability to understand 3D geometry, it introduces the afforementioned bottlenecks due to the point operations. 

%\textit{Given the cubic memory demands of voxel grids and the sparsity of points in 3D space, the Point-Voxel convolution is adapted for smaller neighborhoods, similar to PointNet++.} Nevertheless, PVCNN++ still encounters similar bottlenecks due to the point operations.

Point-Voxel CNNs \cite{pvcnn} introduce a hybrid convolution approach, combining a high-resolution point branch with a low-resolution voxel branch to aggregate neighborhood information. In \cite{pvcnn} the authors propose two network architectures. The first, PVCNN, employs a series of Point-Voxel layers to extract point features. The second variant, PVCNN++, inspired by PointNet++, also utilizes point operations. After each voxel convolution, points are sampled and grouped into neighborhoods, and collective features are extracted for the neighborhood centroids, creating a point encoder with a decreasing number of points. While this approach enhances the network's ability to understand 3D geometry, it introduces the aforementioned runtime bottlenecks due to the point operations.

Sparse Voxel Convolution models \cite{SparseConvNet, SECOND, MinkowskiEngine} address the cubic memory requirements of dense voxel grids, thereby enabling higher voxel densities while preserving fast execution times. In SPVNAS \cite{spvnas} the authors propose the use of Sparse Point-Voxel Convolutions, creating an effective module that combines the efficiency of the sparse voxel convolution and the high fidelity features that are propagated through the point branch.

\subsection{3D Shape Generation} %Point Cloud

3D shape generation involves creating synthetic models, most commonly represented as Point Clouds or Meshes. Early works primarly utilized Autoencoder architectures \cite{SetVAE, FoldingNet}, Generative Adversarial Networks (GANs) \cite{achlioptas18a, valsesia2018learning, 9009495} and Gradient Fields \cite{PointGT}. PointFlow \cite{PointFlow} proposed a probabilistic framework, utilizing continuous normalizing flows, to learn a two-level hierarchy of distributions: a distribution of shapes and of points given a shape. Subsequent flow-based research includes \cite{klokov20eccv, softflow}. 

\cite{DPM} introduces a diffusion-based pipeline where points are projected into a latent space representation using a PointNet-like network, followed by a neural network that gradually denoises the latent space representations, resulting in novel shapes.
In PVD \cite{PVD} the authors study a diffusion network that operates directly on the point space. The diffusion U-Net is based on the architecture of PVCNN++ \cite{pvcnn}. PVD is a pivotal work, as it has enabled the creation of more complex generative pipelines. In \cite{LION} the authors design a latent space diffusion pipeline, where the original point cloud is projected to a latent space and a PVD U-Net is responsible for point cloud denoising. In \cite{PSF} the authors retrain a PVD U-Net by optimizing the curvy trajectory of the diffusion denoising process into a straight path. Furthermore, they propose a distillation process to shorten this straigh path into one step, enabling a fast generation pipeline. 

However, the point operations in the PVD U-Net limit both the runtime of the network and its scalability. Network hyperparameters, such as the size of the point neighborhoods, are highly affected by the total number of points. Consequently, a network architecture cannot be applied to higher density point clouds without significant architectural changes. The same issue applies to attention operations in the point space, as the size of the attention matrix is influenced by the number of points. 

Recent advancements in 3D shape generation also include the generation of 3D meshes, rather than point clouds, utilizing mesh diffusion techniques \cite{MeshDiffusion, DMESH} or improving the quality of generated shapes through refining modules \cite{lyu2022a}.

\section{Sparse Point-Voxel Diffusion}

\subsection{Denoising Diffusion Probabilistic Models}

\begin{figure}
    \centering
    \includegraphics[width=3.3in]{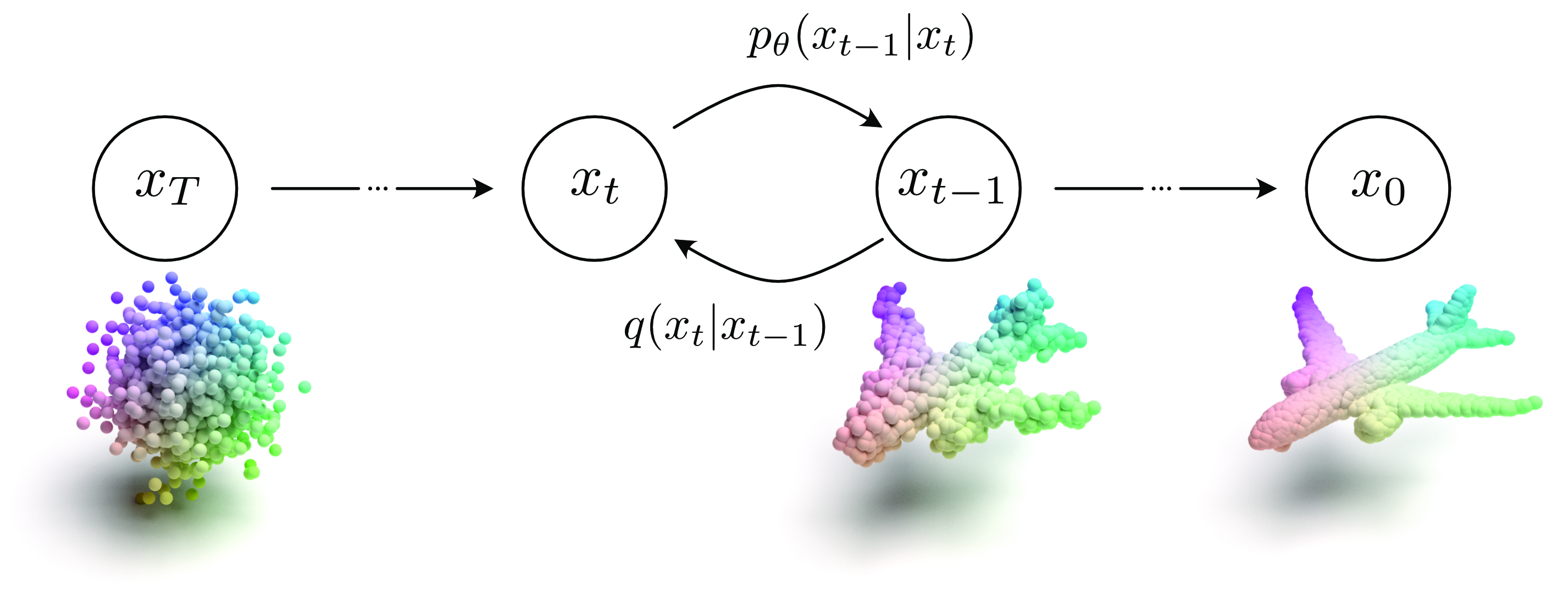}
    \caption{Illustration of the forward and reverse diffusion processes in DDPM. Initially, a clean shape 
  is progressively noisified through the forward diffusion process, resulting in increasingly noisy samples up to $\mathbf{x}_{T}$. These samples, generated via a predefined noise schedule, are utilized during the training phase. The reverse process, indicated by the arrows, involves a neural network tasked with estimating the inverse noise distribution to progressively denoise the samples, eventually reconstructing a clean shape $\mathbf{x}_{0}$.}
    \label{fig:Diffusion Pipeline}
\end{figure}

Denoising Diffusion Probabilistic Models (DDPM) are a class of generative models inspired by thermodynamics \cite{sohl2015deep}, where novel shapes are generated by progressively denoising samples originating from a Gaussian distribution. The process is illustrated in Figure \ref{fig:Diffusion Pipeline}. 

During the forward diffusion process, a sample from the original data distribution, denoted as $\mathbf{x}_{0}$, is progressively corrupted by adding Gaussian noise according to a predefined variance schedule $\beta_1, \ldots, \beta_T$. This  results in a series of progressively noisier samples $\mathbf{x}_{1}, ..., \mathbf{x}_{T}$, of the same dimensionality as  $\mathbf{x}_{0}$. This process can be formulated by stating that the approximate posterior $q(\mathbf{x}_{0:T})$ is defined as a Markov chain. 

\begin{equation}
\label{eq:forwardDiffusion}
\begin{aligned}
q(\mathbf{x}_{0:T}) &= q(\mathbf{x}_0)\prod_{t=1}^{T} q(\mathbf{x}_t | \mathbf{x}_{t-1})\\[0.5em]
q(\mathbf{x}_t | \mathbf{x}_{t-1}) &:= \mathcal{N}(\sqrt{1 - \beta_t} \ \mathbf{x}_{t-1}, \beta_t \ \mathbf{I})
\end{aligned}
\end{equation}

Furthermore, by setting $\alpha_t = 1 - \beta_t$, $\overline{\alpha}_t = \prod_{s=0}^{t}\alpha_s$ and applying reparameterization \cite{reparameterizationTrick} on equation \ref{eq:forwardDiffusion}, a closed-form equation is derived, allowing for direct computation of the sample at timestep $t$: 

% \begin{equation}
% \label{eq:closeform}
% \begin{aligned}
% \mathbf{x}_t &\sim \mathcal{N}(\sqrt{\overline{\alpha}_t} \ \textbf{x}_0, (1 - \overline{\alpha}_t) \ \textbf{I})
% \end{aligned}
% \end{equation}

% or 
% \begin{equation}
% \label{eq:closeform2}
% \begin{aligned}
% \mathbf{x}_t &= \sqrt{\overline{\alpha}_t} \ \textbf{x}_0 + \sqrt{1 - \overline{\alpha}_t} \ \epsilon, \quad \epsilon \sim \mathcal{N}(0, 1)
% \end{aligned}
% \end{equation}

\begin{equation}
\label{eq:closeform}
\begin{aligned}
&\mathbf{x}_t \sim \mathcal{N}(\sqrt{\overline{\alpha}_t} \ \mathbf{x}_0, (1 - \overline{\alpha}_t) \ \mathbf{I}) \\[1em]
\mathbf{x}_t = \ & \sqrt{\overline{\alpha}_t} \ \mathbf{x}_0 + \sqrt{1 - \overline{\alpha}_t} \ \epsilon, \quad \epsilon \sim \mathcal{N}(0, 1)
\end{aligned}
\end{equation}

The reverse diffusion, which is the actual generation phase, involves a neural network predicting the inverse noise distribution, enabling a gradual transition to a clean sample that belongs to the original data distribution. This process is also formulated as a Markov chain with learned Gaussian transitions. 

\begin{equation}
\begin{aligned}
p_\theta(\mathbf{x}_{0:T})  &=  p( \mathbf{x}_T)\prod_{t=1}^{T} p_\theta(\mathbf{x}_{t-1} | \mathbf{x}_t)\\[0.5em]
p_\theta(\mathbf{x}_{t-1} | \mathbf{x}_t) &:= \mathcal{N}(\mu_\theta(\mathbf{x}_{t}, t), {\sigma_t}^2 \, \mathbf{I})
\end{aligned}
\end{equation}

where, $\mu_\theta(\mathbf{x}_{t}, t)$ denotes the shape predicted by the generative model at timestep $t-1$, ${\sigma_t}^2$ is the variance at timestep $t$; the specifics of how this variance is determined are discussed later in this section.

\subsubsection*{Training Objective}

The training objective is to optimize the variational lower bound on the negative log likelihood of the data. This can be expressed as:

\begin{equation}
\label{eq:lowerbound}
\mathbb{E} \left[ - \log p_\theta(\mathbf{x}_0) \right] \leq \mathbb{E}_q \left[ - \log \left( \frac{p_\theta(\mathbf{x}_{0:T})}{q(\mathbf{x}_{1:T} | \mathbf{x}_0)} \right) \right]
\end{equation}

where $q(\mathbf{x}_{1:T} | \mathbf{x}_0)$ is the true posterior, for each timestep $1,...,T$ and $p_\theta(\mathbf{x}_{0:T})$ is the model's approximation. 

Since the noise at timestep $t$ can be computed in closed form by equation \ref{eq:closeform}, the network can be trained effectively by optimizing random terms of variational lower bound \ref{eq:lowerbound}. By following the analysis in \cite{DDPM} the training objective can be simplified to:

\begin{equation}
\begin{aligned}
\mathcal{L} = || \epsilon - \epsilon_\theta(\textbf{x}_t, t) ||^2,  \quad \epsilon \sim \mathcal{N}(0, 1)
\end{aligned}
\end{equation}

where $\epsilon_\theta(\textbf{x}_t, t)$ is expressed by a neural network and $\epsilon$ is the added noise according to equation \ref{eq:closeform}.

\subsubsection*{Generation Process}

For the generation process, we follow the DDPM generation rule \cite{DDPM}, using equation \ref{eq:DDPMGenRule} for timesteps $T, ..., 1$.

\begin{equation}
\label{eq:DDPMGenRule}
\begin{aligned}
\mathbf{x}_{t-1} = \frac{1}{\sqrt{\alpha_t}} \left( \mathbf{x}_t - \frac{1 - \alpha_t}{\sqrt{1 - \overline{\alpha}_t}} \ \epsilon_\theta(\mathbf{x}_t, t) \right) + \sigma_t \mathbf{z}
\end{aligned}
\end{equation}
where $\mathbf{z} \sim \mathcal{N}(0, 1)$ and $\sigma_t$ can be either $\sqrt{\beta_t}$ or $\sqrt{\frac{1 - \overline{\alpha}_{t-1}} {1 - \overline{\alpha}_t}  \beta_t}$. Our experiments indicate that using the latter value for $\sigma_t$ yields slightly better results without a significant difference. During the last iteration, we do not add random noise; $\sigma_1$ is set to $0$.

Additionally, we explore the implicit generation rule introduced in DDIM \cite{DDIM}, described by equation \ref{eq:DDIMGenRule}. This method introduces a non-Markovian generation process, that leverages the same training procedure as DDPM. The generation process in DDIM follows a deterministic trajectory, allowing for the sampling of fewer timesteps, which results in faster generation times. % enables faster sample generation by reducing the number of timesteps required, all while using the same pretrained noise prediction model from DDPM.

\begin{equation}
\label{eq:DDIMGenRule}
\begin{aligned}
\mathbf{x}_t &= \sqrt{\overline{\alpha}_{t-1}} 
\left( \frac{\mathbf{x}_t - \sqrt{1 - \overline{\alpha}_t} \ \epsilon_\theta(\mathbf{x}_t, t)}{\sqrt{\overline{\alpha}_t}} \right) \\[0.5em]
&\quad + \sqrt{1 - \overline{\alpha}_{t-1}} \ \epsilon_\theta(\mathbf{x}_t, t)
\end{aligned}
\end{equation}

\subsection{Sparse Point-Voxel Models}

\begin{figure*}
    \centering
    \includegraphics[width=7in]{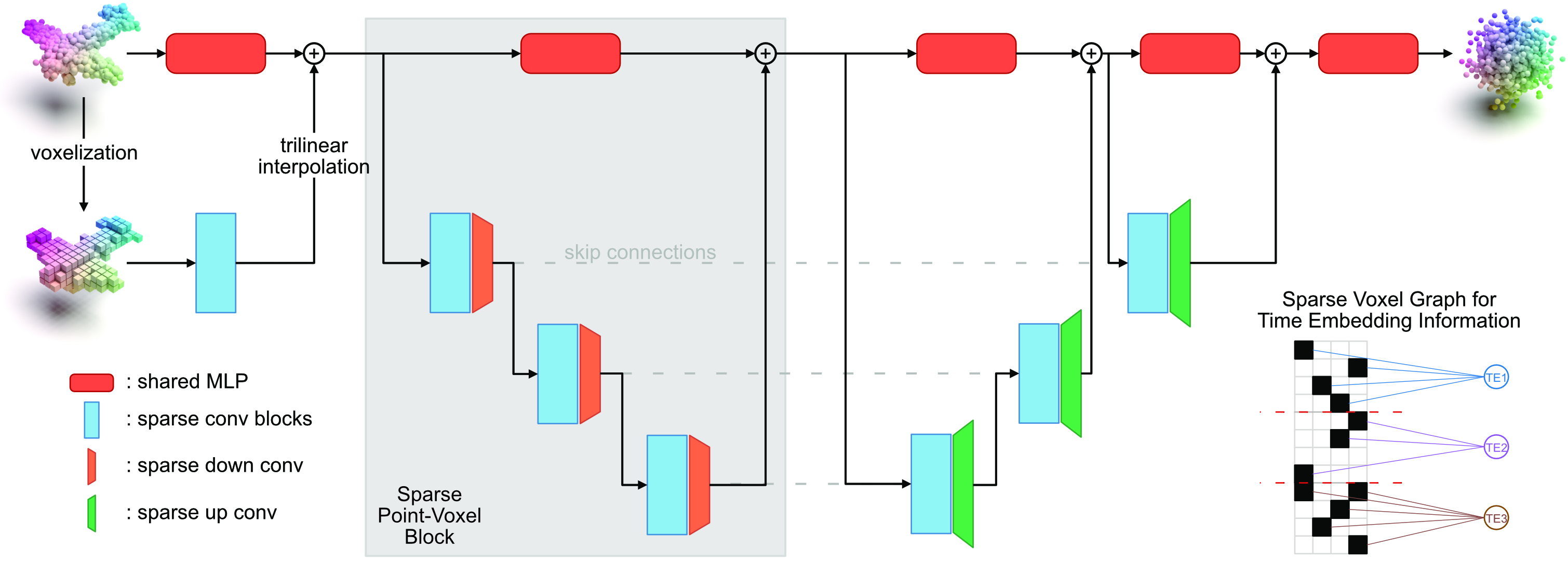}
    \caption{Example architecture of the Sparse Point-Voxel U-Net. The initial point clouds are voxelized and sparse convolutions extract features incorporating neighborhood information. These features are propagated back to the point representation and are merged with the point features, extracted through shared-MLPs. This dual branch architecture is called Sparse Point-Voxel Block (SPVBlock). Note that, as shown, the voxel computations at each SPVBlock may vary, and the point branches in the encoder and decoder do not need to be symmetric. Additionally, we illustrate how sparse voxels and time embeddings can be linked as graph nodes to efficiently handle the varying number of sparse voxels in each point cloud in a batch.}
    \label{fig:NetworkArchitecture}
\end{figure*}

Based on \cite{spvnas}, we propose a novel dual-branch architecture that combines the high fidelity of point representations with the effectiveness of voxel convolutions to extract features, leveraging neighborhood information. Given the complexity and space required to illustrate the entire model, we find it beneficial to present an example architecture that highlights all the key components of the network, as shown in figure \ref{fig:NetworkArchitecture}.

Our model is structured around a series of Sparse Point-Voxel Blocks (SPVBlocks), where both the input and the output are point representations of shape $B$ x $N$ x $F$, with $B$ representing the batch dimension, 
$N$ the number of points per point cloud, and 
$F$ the feature dimension. Initially, the input point cloud is voxelized into a sparse grid which undergoes processing through a combination of Residual Blocks, attention blocks, and either downsampling or upsampling convolution layers. Once processed, the final voxel features are projected back to the original points via trilinear interpolation and then added to the point features, which have been refined through an MLP.

In addition to the SPVBlocks, the combined blocks of the voxel branch form a U-Net network \cite{unet_publication}. The architecture of this model follows the design of the DDPM U-Net \cite{DDPM}, with appropriate adaptations for the sparse voxel domain. Further details are provided in the appendix.

%A key challenge in this sparse voxel approach is incorporating the time embedding information into the network architecture, as there is no constant number of voxels per point cloud in a batch.
A key challenge in this sparse voxel approach is incorporating the time embedding information into the network architecture. The time embedding represents the current step in the denoising process. To integrate the time embedding with the voxel features $F$, we project it to a $scale$ and a $shift$ through the use of a multi-layer perceptron (MLP). 

\begin{figure}
    \centering
    \includegraphics[width=0.95\linewidth]{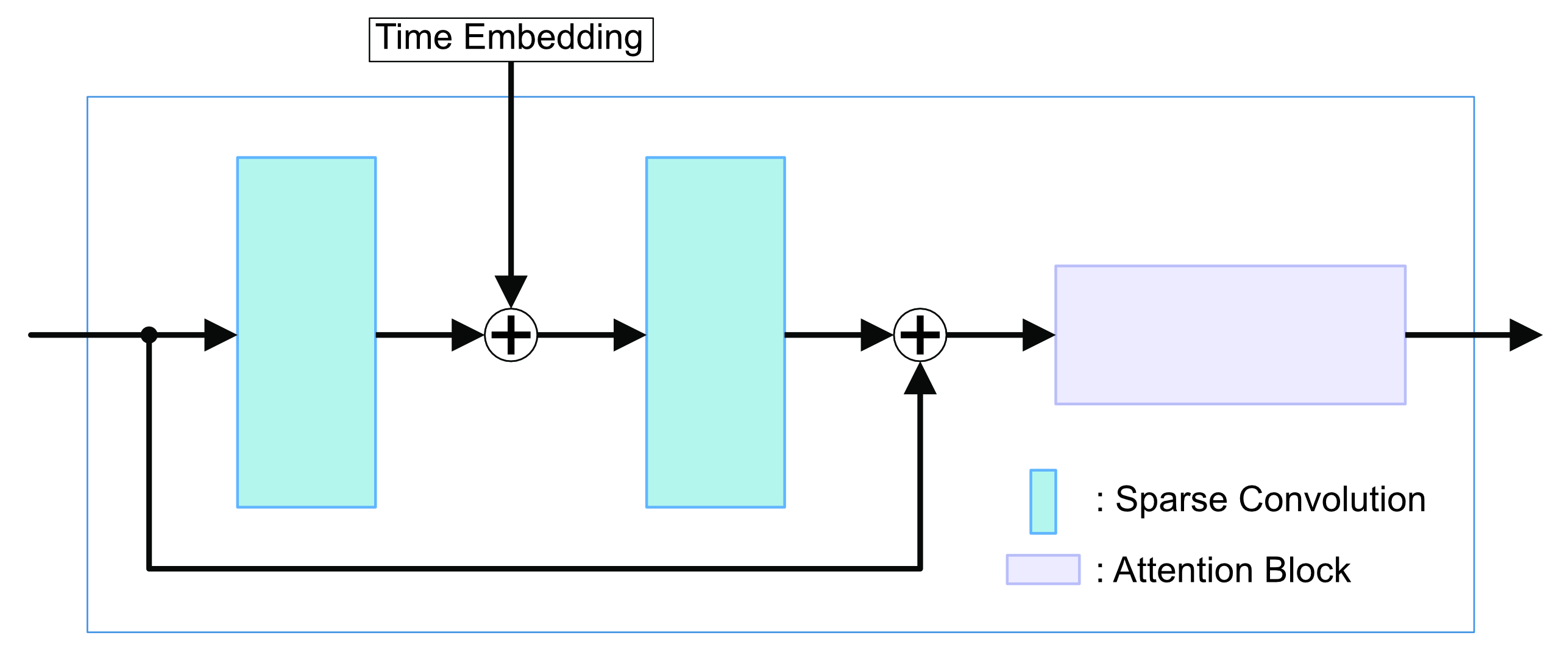}
    \caption{Illustration of a Sparse Residual Convolutional Block. Time embedding information is integrated into the voxel features between two successive convolutional blocks. An optional attention block can further process the voxel features to incorporate global shape information.}
    \label{fig:ConvBlock}
\end{figure}

\begin{equation}
    F' = scale * F + shift
\end{equation}

However, the sparse voxels are stored sequentially for the entire batch, with their numbers varying among the individual point clouds. Furthermore, not all point clouds share the same time embedding, especially during training. This raises the need for an algorithm that would link each sparse voxel with the correct $scale$ and $shift$.

To avoid an iterative process, we implement a novel approach where each sparse voxel and each time embedding are represented as graph nodes and are connected accordingly as illustrated in Figure \ref{fig:NetworkArchitecture}. For our implementation, we utilize PytorchGeometric \cite{PytorchGeometric}, a framework designed for efficient graph processing. Additionally, the tensor storing the sparse voxels and graph nodes follows the same structure, allowing this transition without any computational overhead. This graph-based approach significantly boosts network execution and enables efficient batch processing. 
In figure \ref{fig:ConvBlock} we illustrate the design of a sparse convolutional block. Time embedding information is incorporated between two successive convolutional blocks. The attention block is an optional component.

Finally, it is important to mention that the proposed architecture can process various point densities without any architectural changes, since the point operations are limited to the Shared-MLPs.

\section{Experiments}

\subsection{Unconditional Shape Generation}
\label{sec:UnconditionalGeneration}

\begin{figure*}
    \centering
    \includegraphics[width=0.99\linewidth]{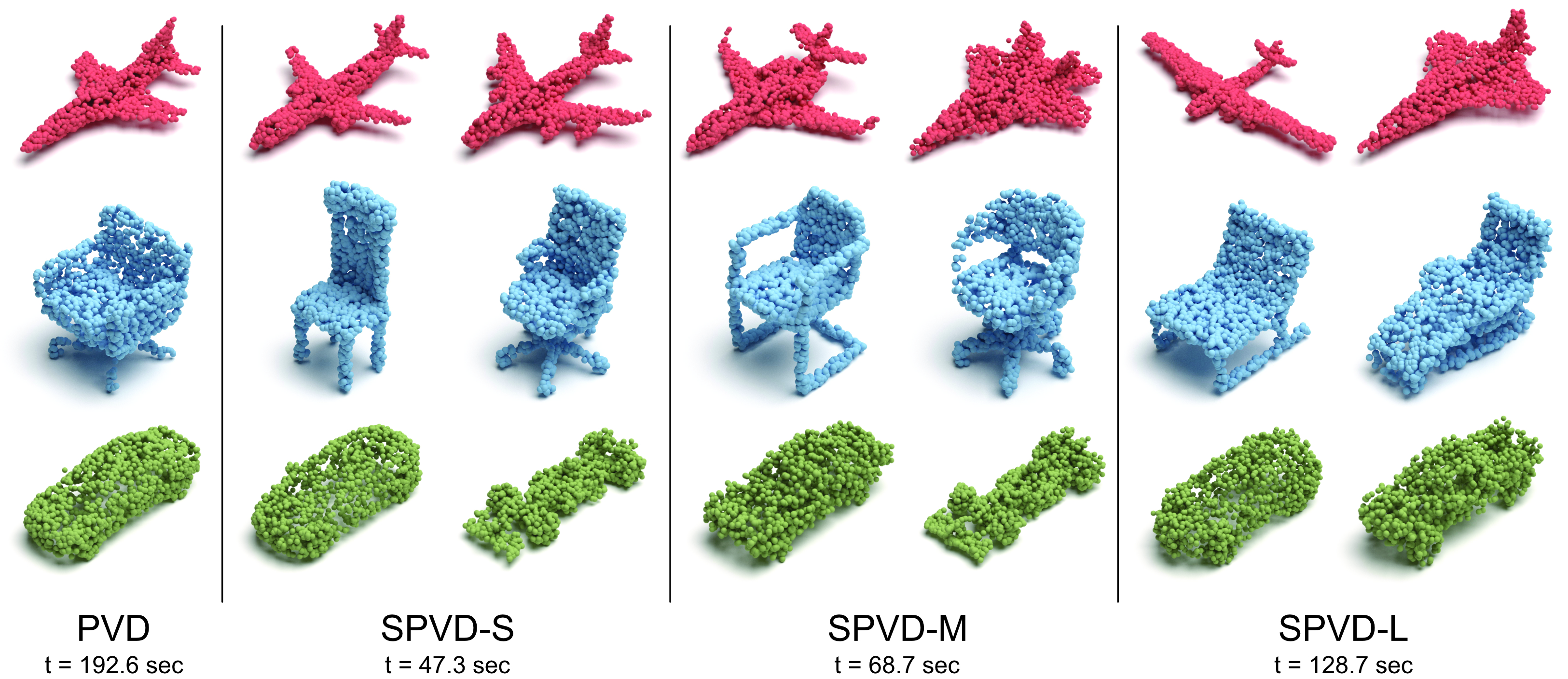}
    \caption{Results of unconditional generation using our three model variants compared to PVD \cite{PVD}. While all models produce high-quality point clouds, our largest models can generate more unique shapes with coarser features, whereas PVD has lower shape diversity. For each model we report the generation time of a batch with 32 samples.}
    \label{fig:comp_eval}
\end{figure*}

\begin{table}[t]
\begin{center}
\caption{Comparative evaluations on shape generation using the 1-NN metrics for ShapeNet Airplane, Chair, and Car categories, employing Chamfer Distance (CD) and Earth Mover Distance (EMD) as distance metrics. Lower ($\downarrow$) scores indicate better generation quality and shape diversity.}

\label{tab:GenResults}
    \small
    \setlength{\tabcolsep}{4pt}
    \begin{tabular}{lcccccc}
        \toprule
        & \multicolumn{2}{c}{Airplane} & \multicolumn{2}{c}{Chair} & \multicolumn{2}{c}{Car} \\
        \cmidrule(lr){2-3} \cmidrule(lr){4-5} \cmidrule(lr){6-7}
        & CD & EMD & CD & EMD & CD & EMD \\
        \midrule
        %\multicolumn{7}{c}{\textit{Non Diffusion Methods}} \\
        %\midrule
r-GAN~\cite{achlioptas18a}         & 98.40 & 96.79  &  83.69 & 99.70  &  94.46 & 99.01 \\
l-GAN (CD)~\cite{achlioptas18a}    & 87.30 & 93.95  &  68.58 & 83.84  &  66.49 & 88.78 \\
l-GAN (EMD)~\cite{achlioptas18a}   & 89.49 & 76.91  &  71.90 & 64.65  &  71.16 & 66.19 \\
Shape-GF~\cite{ShapeGF}            & 80.00 & 76.17  &  68.96 & 65.48  &  63.20 & 56.53 \\
PointFlow~\cite{PointFlow}         & 75.68 & 70.74  &  62.84 & 60.57  &  58.10 & 56.25 \\
SoftFlow~\cite{softflow}    & 76.05 & 65.80  &  59.21 & 60.05  &  64.77 & 60.09 \\
DPF-Net~\cite{klokov20eccv}        & 75.18 & 65.55  &  62.00 & 58.53  &  62.35 & 54.48 \\

\midrule
SPVD-S (\textit{ours}) & \textbf{73.82} & \textbf{64.56} & \textbf{57.10} & \textbf{55.97} & \textbf{56.39}& \textbf{53.83} \\
\midrule
\midrule
   %\multicolumn{7}{c}{\textit{Diffusion Methods}} \\
   %    \midrule
DPM~\cite{DPM}                     & 76.42 & 86.91  &  60.05 & 74.77  &  68.89 & 79.79 \\
PVD~\cite{PVD}                     & 73.82 & 64.81  &  56.26 & 53.32  & \textbf{54.55} & 53.83 \\
%\midrule
%PSF~\cite{PSF}                     & \textbf{71.11} & \textbf{61.09}  &  58.92 & 54.45  &  \underline{57.19} & 56.07 \\
       \midrule
       %\midrule
%LION~\cite{LION}                   & 67.41 & 61.23  &  53.70 & 52.34  &  53.41 & 51.14 \\

        %\midrule
SPVD-M (\textit{ours})               &  73.95     & 63.08 &  56.11  &  57.10  &    70.88    &   52.98    \\
SPVD-L (\textit{ours})               &  \textbf{73.21}  &  \textbf{61.97}      &  \textbf{55.36}  &  \textbf{52.56}  &   70.74     &  \textbf{52.41}    \\
        \bottomrule
    \end{tabular}
\end{center}
\end{table}

For comparative evaluations, we demonstrate our network's results on the ShapeNet \cite{ShapeNet} Airplane, Chair, and Car categories, following the paradigm of previous works \cite{achlioptas18a, softflow, klokov20eccv, DPM, ShapeGF, PointFlow, PSF, PVD}. To ensure fair comparisons, we use the same dataset and preprocessing methods proposed in PointFlow \cite{PointFlow}. 

Unlike other methods, where the model is trained for 10k epochs and interval checkpoints are separately evaluated, we adopt a more efficient approach due to time and energy required for such extensive training and evaluation. We train our model using a one-cycle learning rate scheduler \cite{oneCycle} for a smaller number of epochs and evaluate only the final checkpoint. For each checkpoint, we conduct three evaluation tests and report the best results. This approach acknowledges the stochasticity in the evaluation process, recognizing that running multiple evaluations can increase the chances of producing better results. However, rather than testing hundreds of interval checkpoints, we limit our evaluations to three, which strikes a balance between thoroughness and efficiency.

Evaluation results are presented in Table \ref{tab:GenResults}. The 1-NN metric \cite{PointFlow} is used to evaluate the performance of different methods, while Chamfer Distance (CD) and Earth Mover Distance (EMD) are employed as distance metrics. Lower  scores indicate better generation quality and shape diversity. We observe that the smallest variant of SPVD outperforms non diffusion methods while producing results equivalent to PVD. Furthermore, SPVD-L variant achieves state-of-the-art results across all methods while still maintaining a faster runtime that PVD. Qualitative results of our networks and PVD are presented in Figure \ref{fig:comp_eval}, along with the generation time for a batch of 32 samples. Details of the network variants and additional evaluation metrics are provided in the Appendix.

\subsection{Conditional Generation}

It is important for a general-purpose generative model, akin to the novel generative models in image synthesis, to be capable of generating various objects from a wide range of categories. To test our model's ability to generalize across multiple categories, we train it on all categories of ShapeNet. To select the object category for generation, we use a conditional class embedding, which is incorporated into the pipeline by adding it to the time embedding.

\begin{figure*}
    \centering
    \includegraphics[width=0.95\linewidth]{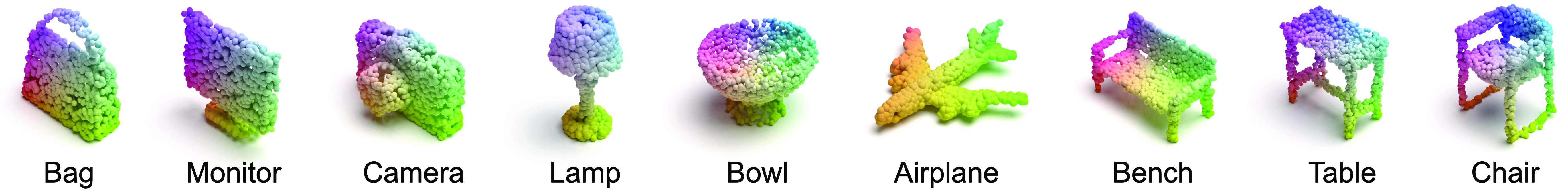}
    \caption{Point clouds generated using the conditional SPVD-L model trained on all categories of ShapeNet. The use of the conditional embedding allows us to specify the class of the generated objects. Our model demonstrates its scalability by generating clean shapes across various categories. }
    \label{fig:ConditionalGeneration}
\end{figure*}

In figure \ref{fig:ConditionalGeneration}, we present generation results of objects from different categories. The results indicate that the model succeeds in generating shapes from diverse categories, demonstrating that the proposed backbone can scale to accommodate more data.

% \begin{table}[t]
% \begin{center}
% \caption{Comparative evaluation of the Conditional Model as well as the unconditional model with the DDIM scheduler against the DDPM conditional versions.  
% Lower ($\downarrow$) scores indicate better generation quality and shape diversity.}
% \label{tab:CondGenResults}
%     \small
%     \setlength{\tabcolsep}{4pt}
%     \begin{tabular}{lcccccc}
%         \toprule
%         & \multicolumn{2}{c}{Airplane} & \multicolumn{2}{c}{Chair} & \multicolumn{2}{c}{Car} \\
%         \cmidrule(lr){2-3} \cmidrule(lr){4-5} \cmidrule(lr){6-7}
%         & CD & EMD & CD & EMD & CD & EMD \\
%         %\midrule
%         %\multicolumn{7}{c}{\textit{Non Diffusion Methods}} \\
%         %\midrule
% %DPM~\cite{DPM} & 76.42 & 86.91  &  60.05 & 74.77  &  68.89 & 79.79 \\
% %PVD~\cite{PVD} & 73.82 & 64.81  &  56.26 & 53.32  & 54.55 & 53.83 \\
%        %\midrule
% %LION~\cite{LION}                   & 67.41 & 61.23  &  53.70 & 52.34  &  53.41 & 51.14 \\

%         \midrule
% SPVD     &  73.95  &  63.08 &  56.11  &  57.10  &  70.73  & 56.96    \\
% SPVD-L   &  73.21  &  61.97 &  55.36  &  52.56  &  70.74  & 52.41    \\
%         \midrule
% SPVD-L \textit{Cond} &  99.62  &  99.01 &  60.19 & 53.39  & 91.05 & 64.48 \\
%         \midrule
% SPVD-L \textit{DDIM} &  73.21  &  61.97 &  60.19 & 53.39  & 91.05 & 64.48 \\
%         \bottomrule
%     \end{tabular}
% \end{center}
% \end{table}

\subsection{Implicit Generation}

\begin{figure}
    \centering
    \includegraphics[width=0.99\linewidth]{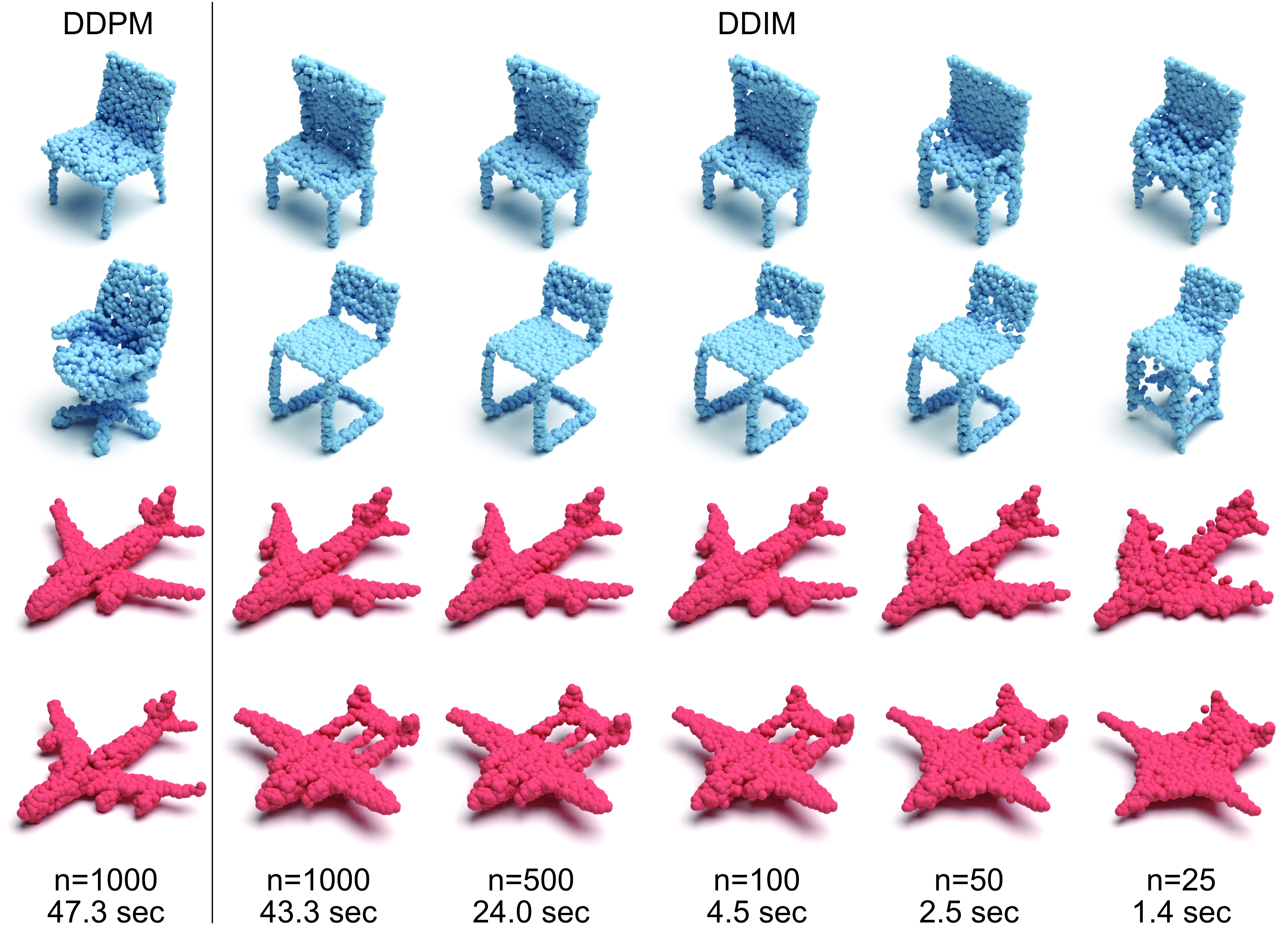}
    \caption{Comparison of probabilistic (DDPM) and implicit (DDIM) generation. Each row displays shapes generated from the same initial random noise. For each shape, the number of sampling steps and the generation time for a batch of 32 point clouds are reported. The implicit model consistently converges to the same shape, demonstrating that DDIMs follow a deterministic trajectory during the denoising process. Additionally, the generative model can produce high-quality samples even with 100 sampling steps, reducing the initial sampling time to one-tenth.}
    \label{fig:implicitGeneration}
\end{figure}

\begin{table}[t]
\begin{center}
\caption{1-NN metric and generation times for the SPVD variants using the DDIM generation rule with 100 and 1000 steps. The DDPM metrics are also included for comparison.}

\label{tab:ImplicitGenResults}
    \small
    \setlength{\tabcolsep}{4pt}
    \begin{tabular}{l|c|cccccc}
        \toprule
        \multirow{3}{*}{\shortstack{DDIM\\steps}} & \multirow{3}{*}{\shortstack{Gen\\(sec)}} & \multicolumn{2}{c}{Airplane} & \multicolumn{2}{c}{Chair} & \multicolumn{2}{c}{Car} \\
        \cmidrule(lr){3-4} \cmidrule(lr){5-6} \cmidrule(lr){7-8}
        & & CD & EMD & CD & EMD & CD & EMD \\
        \midrule

  \multicolumn{8}{c}{SPVD-S - \textbf{23M} parameters} \\
\midrule
100 & 4.5 & 85.80 & 65.92 & 64.73 & 60.50 & 66.19 & 57.81 \\
1000 & 43.3 & 75.18 & 65.18 & 61.63 & 59.74 & 60.36 & 60.22 \\
\midrule
DDPM & 47.3 & 73.82 & 64.56 & 57.10 & 55.97 & 56.39 & 53.83\\
\midrule
  \multicolumn{8}{c}{SPVD-M - \textbf{33M} parameters} \\
\midrule
%    | Time |    Airplane   |     Chair     |      Car 
100  &  6.7 & 84.19 & 75.80 & 75.37 & 77.26 & 79.97 & 77.41 \\
1000 & 68.4 & 84.56 & 79.75 & 78.32 & 79.07 & 87.21 & 83.38 \\
\midrule
DDPM & 68.7 & 73.95 & 63.08 & 56.11 & 57.10 & 70.88 & 52.98 \\
\midrule
 \multicolumn{8}{c}{SPVD-L - \textbf{88M} parameters} \\
\midrule
%    | Time |    Airplane   |     Chair     |      Car 
100  & 12.8 & 84.07 & 71.85 & 69.48 & 71.48 & 78.12 & 75.00 \\
1000 & 127.6 & 80.00 & 78.02 & 70.77 & 71.97 & 80.25 & 76.42 \\
\midrule
DDPM & 128.7 & 73.21 & 61.97 & 55.36  & 52.56  & 70.74 & 52.41 \\

        \bottomrule
    \end{tabular}
\end{center}
\end{table}

In this section, we evaluate the results of our network when using the DDIM rule for shape generation. Implicit generation, as proposed by \cite{DDIM}, suggests that the generation process follows a deterministic trajectory and allows for generation with fewer sampling steps. The aim of this experiment is to decrease the point cloud generation time while observing the effect on generation quality.

% The network used in our experiments is the SPVD-S variant. We do not perform any new training; instead, we use the weights proposed for unconditional shape generation in section \ref{sec:UnconditionalGeneration}.

In Figure \ref{fig:implicitGeneration}, we illustrate the generation results for various sampling steps, all starting from the same random noise. We present results for the chair and airplane categories, chosen for their distinguishable features. The model used in this experiment is the SPVD-S variant, trained for unconditional generation. 
It is evident that our model can generate high-quality shapes with just 100 steps, equivalent to one-tenth of the initial generation time. This speed-up could enable a range of applications where execution speed is prioritized over absolute accuracy.

In Table \ref{tab:ImplicitGenResults} we present the 1-NN metric for the DDIM generation rule using 100 and 1000 generation steps along with the corresponding runtime.The qualitative results of Figure \ref{fig:implicitGeneration} are quantitatively verified for the SPVD-S variant, demonstrating high generation quality even with just 100 steps. Interestingly, the largest models experience a significant performance drop. This may be due to the larger number of parameters and the lack of random noise during generation, which favors variability.

\subsection{Shape Completion}

We further test our model's ability to complete incomplete shapes by proposing a new task called Part Completion. The input to the model consists of shapes from the PartNet \cite{PartNet} dataset, where random parts of the objects have been removed. More details are provided in the appendix. The task for the model is to reconstruct the missing areas. The key challenges of this task include the variable number of input points, as the selection and number of discarded parts are random, leading to varying input shapes. Additionally, there is no information about the location of the missing points, requiring the model to infer the underlying geometry.

During training, the input points remain constant while only the selected parts for removal are noisified. The model is tasked with estimating the added noise, similar to the generation pipeline. At inference, random noise is concatenated with the input, and the model gradually denoises this new noise to reconstruct the missing parts.

\begin{figure*}
    \centering
    \includegraphics[width=0.99\linewidth]{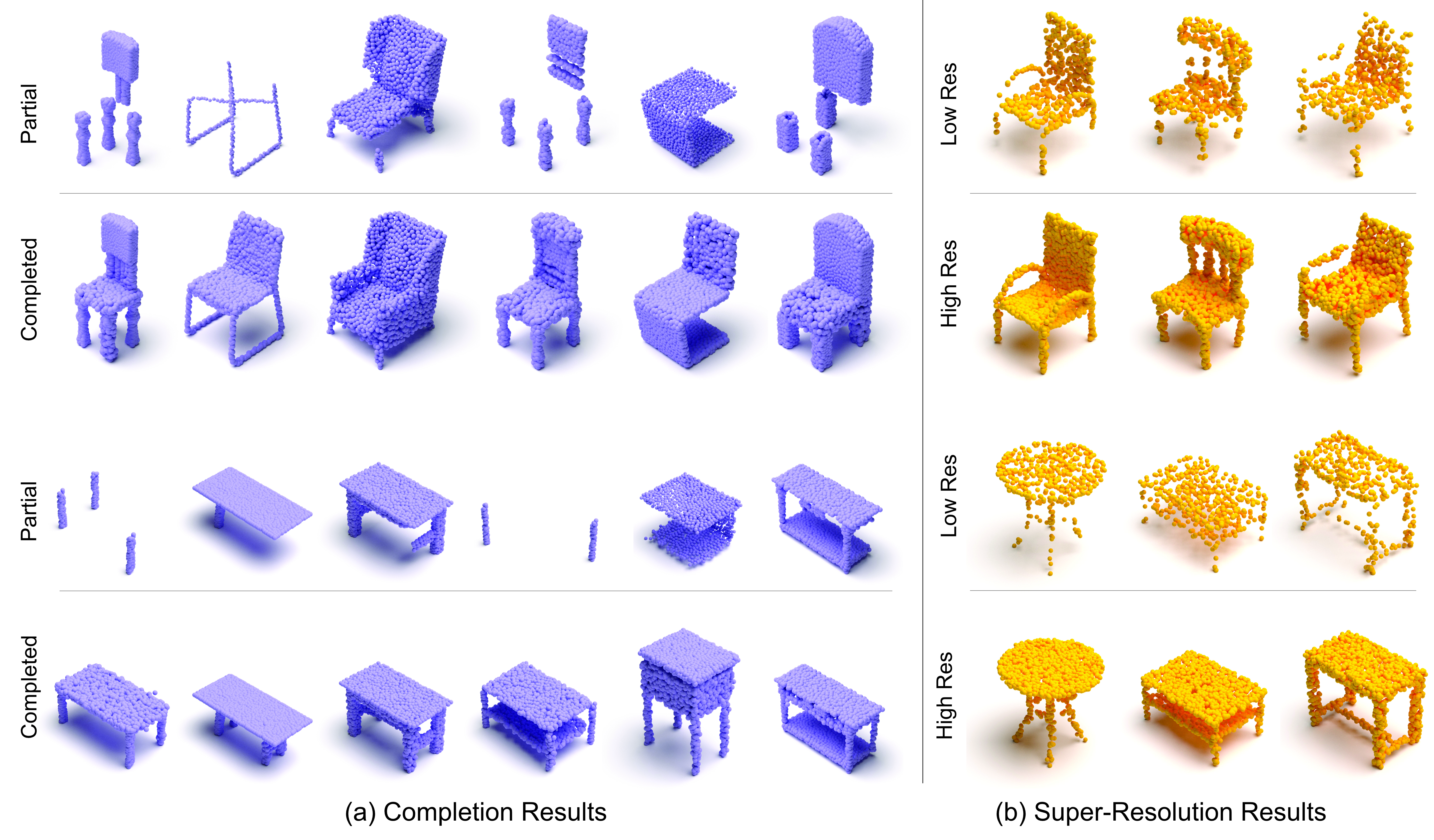}
    \caption{Results of (a) completion and (b) super-resolution networks. (a) The completion network successfully fills in missing parts of the objects without any guidance, predicting only the missing points for input point clouds with varying point counts. (b) The super-resolution network not only increases point density but also adds details to shapes (such as the back of the chair) and fills gaps, like missing points in the chair handles or uneven chair or table legs.}
    \label{fig:CompletionSuperRes}
\end{figure*}

Completion results are presented in Figure \ref{fig:CompletionSuperRes} (a). We showcase results for the chair and table categories, chosen for their distinctive features. The network successfully completes the missing parts of the shapes.

\subsection{Point Cloud Super Resolution}

Super-Resolution is a well-known task in the image domain, but it has received limited attention in Point Clouds. Structurally, for diffusion models, super-resolution is similar to completion, with the primary difference being that in super-resolution, random points are noisified, while in completion, specific parts. For our experiment, we use a point cloud with 512 points as input and generate an output with 2048 points.

The results of the super-resolution task are shown in Figure \ref{fig:CompletionSuperRes} (b). We notice that the low sampling rate of the input results in many missing details, which can alter the shape, such as uneven chair legs or missing parts in the handles and back. The model not only increases the resolution of the shapes but also fills in the missing information.

\section{Conclusion and Future Work}

In this paper, we introduced SPVD, a novel diffusion U-Net architecture that enables efficient and scalable point cloud generation. Our approach achieves state-of-the-art results in point cloud generation with significantly shorter generation times compared to other high-fidelity diffusion models. 
Our experiments demonstrate that SPVD can scale to larger datasets and achieve even faster generation times through implicit generation. 
Additionally, it proves to be a strong candidate for tasks such as point cloud completion and super-resolution.

Fast generative models for 3D shapes are crucial for applications where user experience is just as important as generation quality. SPVD represents a significant step forward in addressing this need.

Future research will focus on developing a latent diffusion pipeline, similar to  \cite{latentDiffusion} in the image domain, where the SPVD model operates within the latent space to denoise latent representations. This pipeline will also facilitate the use of guidance, allowing images or text prompts to influence the generative results. Given the computational efficiency of our model, this latent variant could be extended beyond the ShapeNet to more practical datasets such as Objaverse \cite{objaverse} and Objaverse-XL \cite{objaverseXL}. However, these extensive datasets require filtering and selection of suitable models for training a diffusion model, as well as preprocessing — a separate area for future work.

Another area of future research involves developing novel evaluation metrics that can better distinguish between generation quality and shape diversity. It may also involve identifying distance metrics more suitable than Chamfer and Earth Mover distances. An example from the image domain is the Fréchet Inception Distance (FID) \cite{FID}, which compares feature maps from specific layers of InceptionNet \cite{InceptionNet} rather than the images themselves.

\section*{Acknowledgments}
The present paper has been developed as part of the RRREMAKER project, funded by the European Union 2020 Research and Innovation program under the Marie Sklodowska Curie - RISE grant agreement no 101008060 and the framework of H.F.R.I call “Basic research Financing (Horizontal support of all Sciences)” under the National Recovery and Resilience Plan “Greece 2.0” funded by the European Union -- NextGenerationEU (H.F.R.I. Project Number: 16469.).

{\appendices

\section{Network Desing Details}
In this section, we provide the implementation details of the three network variants.

The SPVD-S variant uses only a single SPVD block, meaning that there is a point representation only at the start and end of the voxel U-Net. The SPVD-M and SPVD-L variants share the same architecture but differ in latent space dimensions. An overview of the architectures is presented in Table \ref{tab:A_NetworkDetails}.

A minor difference between the architectures concerns how the latent space is increased. In the SPVD-S variant, the latent space is increased during the first convolution of each down block. In contrast, for the SPVD-M and SPVD-L variants, the increase occurs during the downsampling convolution, that is the last layer of the block. This approach reduces the parameter count, allowing for larger feature dimensions.

Our experiments show that training larger models with the architecture of the SPVD-S variant did not yield better results. We believe that the point skip connections in the larger variants facilitate the propagation of gradients, enabling the successful training of larger architectures.

\begin{table*}[]
    \centering
    \renewcommand{\arraystretch}{1.2}
    \caption{Architectural details of the proposed architectures. The first row lists the layers of the voxel U-Net architecture. For each architecture, we report the layers used, the feature dimensions, whether the output of a block is projected to the point space (indicating the end of an SPVD block and the start of a new one), if there is an upsampling or downsampling convolution in those layers, and if there is a voxel attention block. The symbols used are as follows: \checkmark indicates the presence of the feature in the block, \texttimes indicates that the feature is explicitly set to False, \textbf{-} indicates that the feature is not applicable to the specific layer or is an optional feature that is skipped, and \textbf{na} indicates that the layer does not exist in the architecture, making the value not applicable.}
    \label{tab:A_NetworkDetails}
    \small
    \setlength{\tabcolsep}{4pt}
    \begin{tabularx}{\textwidth}{|c|*{12}{>{\centering\arraybackslash}X|}}
    \hline
     \textbf{Blocks: } & stem & down1 & down2 & down3 & down4 & down5 & mid & up1 & up2 & up3 & up4 & up5 \\
     \hline
     \multicolumn{13}{|c|}{SPVD-S} \\
     \hline
     feature dim & 32 & 32 & 64 & 128 & 256 & na & 256 & 256 & 128 & 64 & 32 & na \\
     project to points & - & - & - & - & - & na & - & - & - & - & \checkmark & na \\
     %downsample & - & \checkmark & \checkmark & \checkmark & \texttimes & na & - & - & - & - & - & na \\
     %upsample & - & - & - & - & - & na & - & \checkmark & \checkmark & \checkmark & \texttimes & na \\
     down/up sample & - & \checkmark & \checkmark & \checkmark & \texttimes & na & - & \checkmark & \checkmark & \checkmark & \texttimes & na \\
     use attention & \texttimes & \texttimes & \texttimes & \texttimes & \checkmark & na & \texttimes & \checkmark & \texttimes & \texttimes & \texttimes & na \\
     \hline
     \multicolumn{13}{|c|}{SPVD-M} \\
     \hline
     feature dim & 32 & 64 & 128 & 192 & 192 & 256 & na & 256 & 192 & 128 & 64 & 32 \\
     project to points & \checkmark & - & - & - & - & \checkmark & na & - & \checkmark & - & - & \checkmark \\
     %downsample & - & \checkmark & \checkmark & \checkmark & \checkmark & \texttimes & na & - & - & - & - & - \\
     %upsample & - & - & - & - & - & - & na & \checkmark & \checkmark & \checkmark & \checkmark & \texttimes \\
     down/up sample & - & \checkmark & \checkmark & \checkmark & \checkmark & \texttimes & na & \checkmark & \checkmark & \checkmark & \checkmark & \texttimes  \\
     use attention & \texttimes & \texttimes & \texttimes & \texttimes & \checkmark & \checkmark & na & \checkmark & \checkmark & \texttimes & \texttimes & \texttimes \\
     \hline
     \multicolumn{13}{|c|}{SPVD-L} \\
     \hline
     feature dim & 64 & 128 & 192 & 256 & 384 & 384 & na & 384 & 256 & 192 & 128 & 64 \\
     project to points & \checkmark & - & - & - & - & \checkmark & na & - & \checkmark & - & - & \checkmark \\
     %downsample & - & \checkmark & \checkmark & \checkmark & \checkmark & \texttimes & na & - & - & - & - & - \\
     %upsample & - & - & - & - & - & - & na & \checkmark & \checkmark & \checkmark & \checkmark & \texttimes \\
     down/up sample & - & \checkmark & \checkmark & \checkmark & \checkmark & \texttimes & na & \checkmark & \checkmark & \checkmark & \checkmark & \texttimes  \\
     use attention & \texttimes & \texttimes & \texttimes & \texttimes & \checkmark & \checkmark & na & \checkmark & \checkmark & \texttimes & \texttimes & \texttimes \\
    \hline
    \end{tabularx}
\end{table*}

\section{Additional generation evaluation metrics}

In this section of the appendix, we provide a comparative evaluation against other methods following \cite{PVD}. The metrics used are Coverage (Cov) and Minimum Matching Distance (MMD). These metrics are complementary: Cov evaluates the diversity of the generated shapes compared to the test set, while MMD measures the quality of the generated shapes.

Although these methods are considered less reliable than the 1-NN metric \cite{PointFlow}, our models still achieve state-of-the-art results, as shown in Table \ref{tab:GenResultsAdditional}.

\begin{table*}[t]
\begin{center}
\caption{Additional generation evaluation metrics.}
\label{tab:GenResultsAdditional}
    \small
    \setlength{\tabcolsep}{4pt}
    \begin{tabular}{lcccc|cccc|cccc}
        \toprule

        & \multicolumn{4}{c}{Airplane} & \multicolumn{4}{c}{Chair} & \multicolumn{4}{c}{Car} \\
        \cmidrule(lr){2-5} \cmidrule(lr){6-9} \cmidrule(lr){10-13}
        
        \multirow{3}{*}{Model} &  \multicolumn{2}{c}{MMD ($\downarrow$)} & \multicolumn{2}{c}{COV ($\%$, $\uparrow$)} &  \multicolumn{2}{c}{MMD ($\downarrow$)} & \multicolumn{2}{c}{COV ($\%$, $\uparrow$)}  & \multicolumn{2}{c}{MMD ($\downarrow$)} & \multicolumn{2}{c}{COV ($\%$, $\uparrow$)}\\

        \cmidrule(lr){2-3} \cmidrule(lr){4-5} \cmidrule(lr){6-7} \cmidrule(lr){8-9} \cmidrule(lr){10-11} \cmidrule(lr){12-13}
        &  CD & EMD & CD & EMD & CD & EMD & CD & EMD & CD & EMD & CD & EMD \\
        \midrule
        %\multicolumn{7}{c}{\textit{Non Diffusion Methods}} \\
        %\midrule
r-GAN~\cite{achlioptas18a}         & 0.4471 & 2.309  &  30.12 & 14.32  &  5.151 & 8.312 & 24.27 & 15.13 & 1.446 & 2.133 & 19.03 & 6.539 \\
l-GAN (CD)~\cite{achlioptas18a}    & 0.3398 & 0.5832 & 38.52 & 21.23 & 2.589 & 2.007 & 41.99 & 29.31 & 1.532 & 1.226 & 38.92 & 23.58 \\
l-GAN (EMD)~\cite{achlioptas18a}   & 0.3967 & 0.4165 & 38.27 & 38.52 & 2.811 & 1.619 & 38.07 & 44.86 & 1.408 & 0.8987 & 37.78 & 45.17 \\
Shape-GF~\cite{ShapeGF}            & 2.703 & 0.6592 & 40.74 & 40.49 & 2.889 & 1.702 & 46.67 & 48.03 & 9.232 & 0.7558 & \textbf{49.43} & 50.28 \\
PointFlow~\cite{PointFlow}         &  0.2243 & 0.3901 & 47.90 & 46.41 & \textbf{2.409} & 1.595 & 42.90 & 50.00 & \textbf{0.9010} & 0.8071 & 46.88 & 50.00 \\
SoftFlow~\cite{softflow}           & 0.2309 & 0.3745 & 46.91 & 47.90 & 2.528 & 1.682 & 41.39 & 47.43 & 1.187 & 0.8594 & 42.90 & 44.60 \\
DPF-Net~\cite{klokov20eccv}        &  0.2642 & 0.4086 & 46.17 & 48.89 & 2.536 & 1.632 & 44.71 & 48.79 & 1.129 & 0.8529 & 45.74 & 49.43 \\

\midrule
SPVD-S (\textit{ours}) & \textbf{0.2281} & \textbf{0.3807} & \textbf{48.64} & \textbf{49.62} & 2.545 & \textbf{1.549} &  \textbf{47.88} & \textbf{52.11} & 0.9155 & \textbf{0.7501} & 45.73 & \textbf{53.40} \\
\midrule
\midrule
   %\multicolumn{7}{c}{\textit{Diffusion Methods}} \\
   %    \midrule
PVD~\cite{PVD}                     & 0.2243 & 0.3803 & \textbf{48.88} & 52.09 & 2.622 & 1.556 & 49.84 & 50.60 & 1.077 & 0.7938 & 41.19 & 50.56 \\

       \midrule

        %\midrule
SPVD (\textit{ours})               &  \textbf{0.2187}   & \textbf{0.3661} &  44.44  &  51.11  &  0.2589  &  1.579 & 48.94 & 51.51 & 0.9947 & 0.7855 & \textbf{46.02} & \textbf{51.98}   \\
SPVD-L (\textit{ours})               &  0.2247  &  0.3705  &  48.46  &  \textbf{53.33}  &  0.2562    &  \textbf{1.521} & \textbf{51.15} & \textbf{52.87} & \textbf{0.9782} & \textbf{0.7495} & 45.45 & 51.70 \\
        \bottomrule
    \end{tabular}
\end{center}
\end{table*}

\section{Part Completion}

In the context of circular economy and recycling initiatives, the repair and reuse of older objects are highly encouraged. Generative models can contribute to this effort by reconstructing missing parts, facilitating the retrieval of replacement parts in databases, or enabling 3D printing solutions. Additionally, 3D model designers could benefit from automated algorithms that either complete their shapes or suggest alternatives for specific parts.

To this aim, we propose a new task called Part Completion. We use PartNet \cite{PartNet}, a 3D model dataset with hierarchical part annotations. We preprocess the models by following the hierarchical structure of their parts and representing it as a tree. Since some parts may be too small, resulting in minimal scan information, we merge all leaves with a low point count into their parent nodes. We then save the resulting point cloud along with their per-point part labels.

During training and evaluation, we set 
$m$ as the minimum number of parts that a partial object should have. We then randomly select a number between 1 and $m$ to determine the parts of the object to discard.
}

% {\appendix[Proof of the Zonklar Equations]
% Use $\backslash${\tt{appendix}} if you have a single appendix:
% Do not use $\backslash${\tt{section}} anymore after $\backslash${\tt{appendix}}, only $\backslash${\tt{section*}}.
% If you have multiple appendixes use $\backslash${\tt{appendices}} then use $\backslash${\tt{section}} to start each appendix.
% You must declare a $\backslash${\tt{section}} before using any $\backslash${\tt{subsection}} or using $\backslash${\tt{label}} ($\backslash${\tt{appendices}} by itself
%  starts a section numbered zero.)}


% \newpage

\bibliographystyle{IEEEtran}
\bibliography{bibliography}

 
\vspace{11pt}

%\bf{If you include a photo:}\vspace{-33pt}
\begin{IEEEbiography}[{\includegraphics[width=1in,height=1.25in,clip,keepaspectratio]{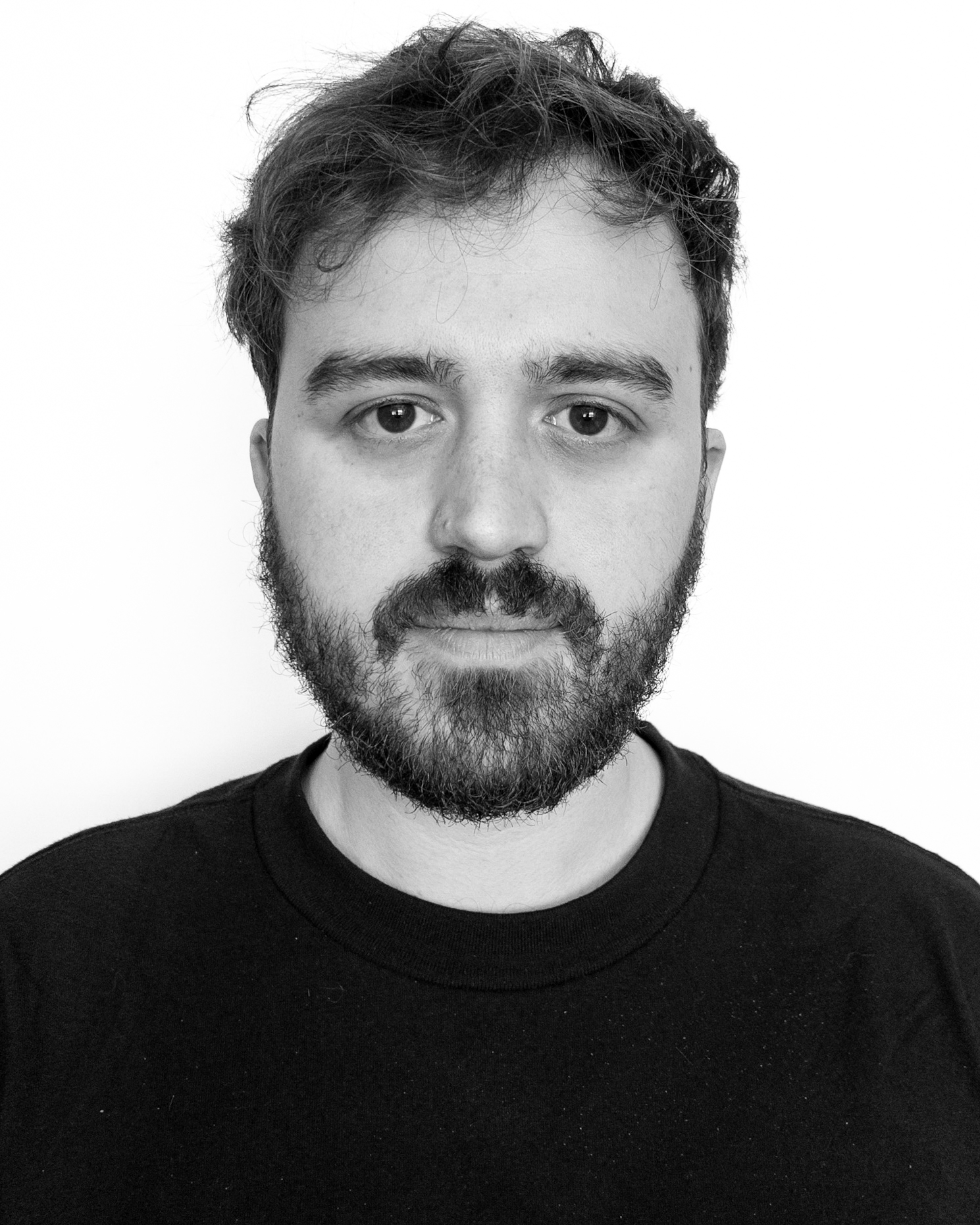}}]{Ioannis Romanelis}
received his Electrical and Computer Engineering diploma in 2021 at the University of Patras. During the same year, he enrolled for a PhD at the same department under the supervision of Professor Konstantinos Moustakas and joined the Visualization and Virtual Reality (VVR) group. His main research interests include computer vision, deep learning, point cloud processing, 3D scene understanding, generative modeling, and explainable AI.
\end{IEEEbiography}

\begin{IEEEbiography}[{\includegraphics[width=1in,height=1.25in,clip,keepaspectratio]{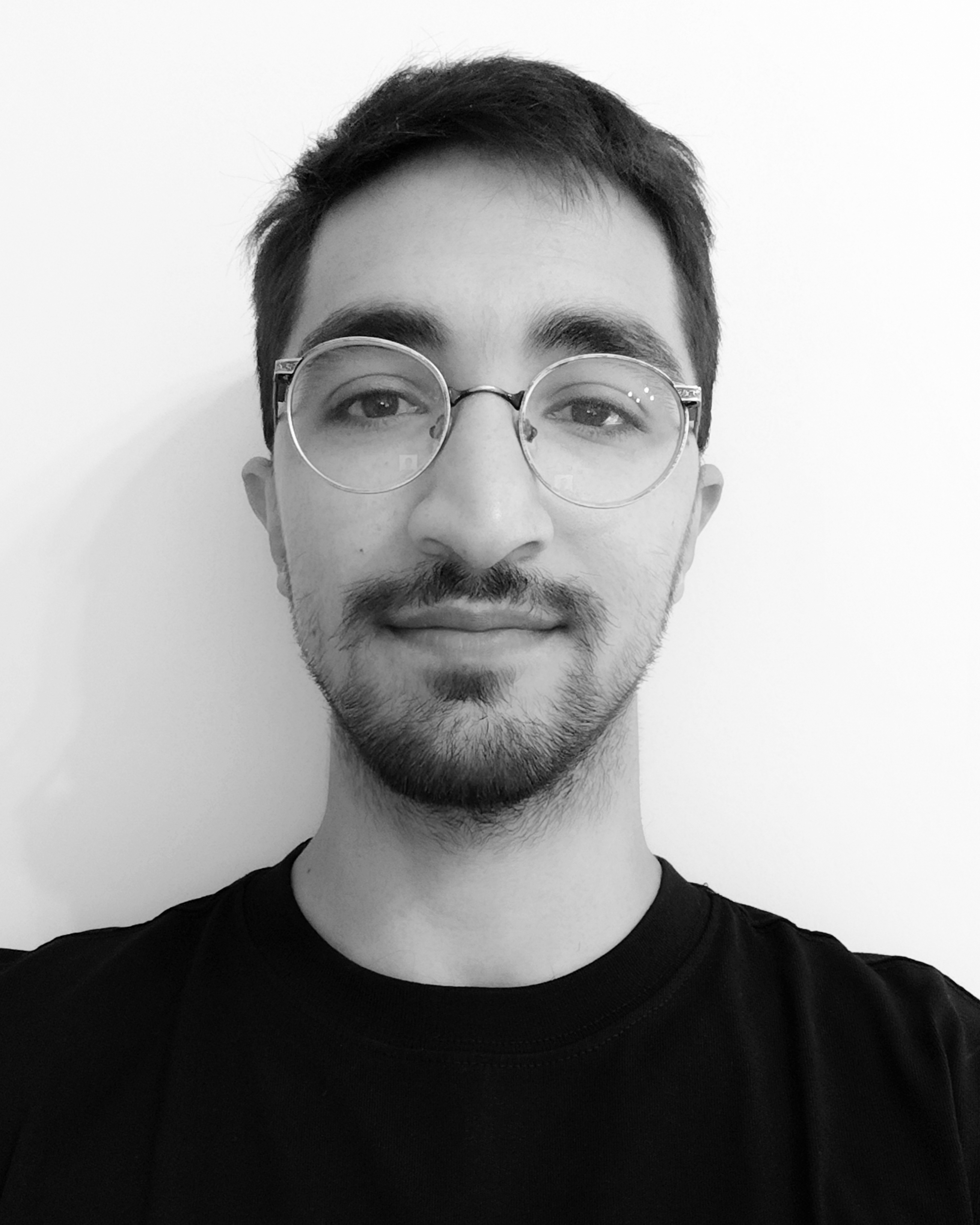}}]{Vlassis Fotis}
received his Electrical Engineering and Computer Technology degree in 2021 at the university of Patras. During the same year he joined the VVR lab as a PhD candidate. His main research interests include (but are not limited to) computer vision, theoretical deep learning, 3D scene understanding and geometry processing.
\end{IEEEbiography}

\begin{IEEEbiography}[{\includegraphics[width=1in,height=1.25in,clip,keepaspectratio]{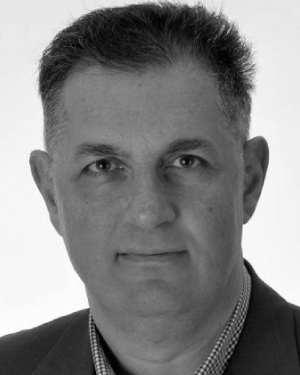}}]{Athanasios Kalogeras}
(Senior Member, IEEE) received the Diploma degree in electrical
engineering and the Ph.D. degree in electrical and computer engineering from the University of Patras, Greece. He has been with the Industrial Systems Institute, ATHENA Research and Innovation Center, since 2000, where he currently holds a position of the Research Director. He has worked as an Adjunct Faculty at the Technological Educational Foundation of Patras. He has been a Collaborating Researcher at the University of Patras, the Computer Technology Institute and Press “Diophantus,” and the private sector. His research interests include cyber-
physical systems, the Industrial IoT, industrial integration and interoperability, and collaborative manufacturing. Application areas include the manufacturing environment, critical infrastructure protection, smart buildings, smart cities, smart energy, circular economy, health, and tourism and culture. He has served as a program committee member for more than 30 conferences and as a reviewer in more than 40 international journals and conferences. He has been a Postgraduate Scholar of the Bodossaki Foundation. He is a member of the Technical Chamber of Greece. He is a Local Editor in Greece of ERCIM News.
\end{IEEEbiography}

\begin{IEEEbiography}[{\includegraphics[width=1in,height=1.25in,clip,keepaspectratio]{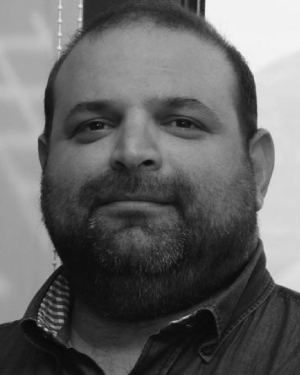}}]{Christos Alexakos}
(Member, IEEE) received the D.Eng. degree in computer engineering and
informatics, the M.Sc. degree in computer science and engineering, and the Ph.D. degree from the Department of Computer Engineering and Informatics, University of Patras, Greece. He was the Technical Manager of the Pattern Recognition Laboratory for ten years and the project manager in four development projects carried out by the laboratory. In 2013, he co-founded InSyBio Ltd., a bioinformatics company that delivers a cloud-based software suite for intelligent analysis of biological big data. He is currently Principal Researcher with the Industrial Systems Institute, ATHENA Research and Innovation Center, Greece. He is collaborating as a Research
Engineer with the Pattern Recognition Laboratory, Department of Computer Engineering and Informatics, University of Patras, and the Computer Technology Institute. He has multi-year experience in development of software applications, both web-based and standalone, including databases applications, GIS, and service-oriented applications. He is a highly experienced programmer and a software architect and has worked as a freelancer in ICT projects for both public and private sector, since 2003. He has published 15 articles in journals, 49 in conference proceedings, and five chapters in books. He also has a submitted patent to USPO. His main research interests include information systems architecture and integration in the fields of enterprise and manufacturing processes, bioinformatics, the IoT, cloud computing, and cloud
manufacturing.
\end{IEEEbiography}

\begin{IEEEbiography}[{\includegraphics[width=1in,height=1.25in,clip,keepaspectratio]{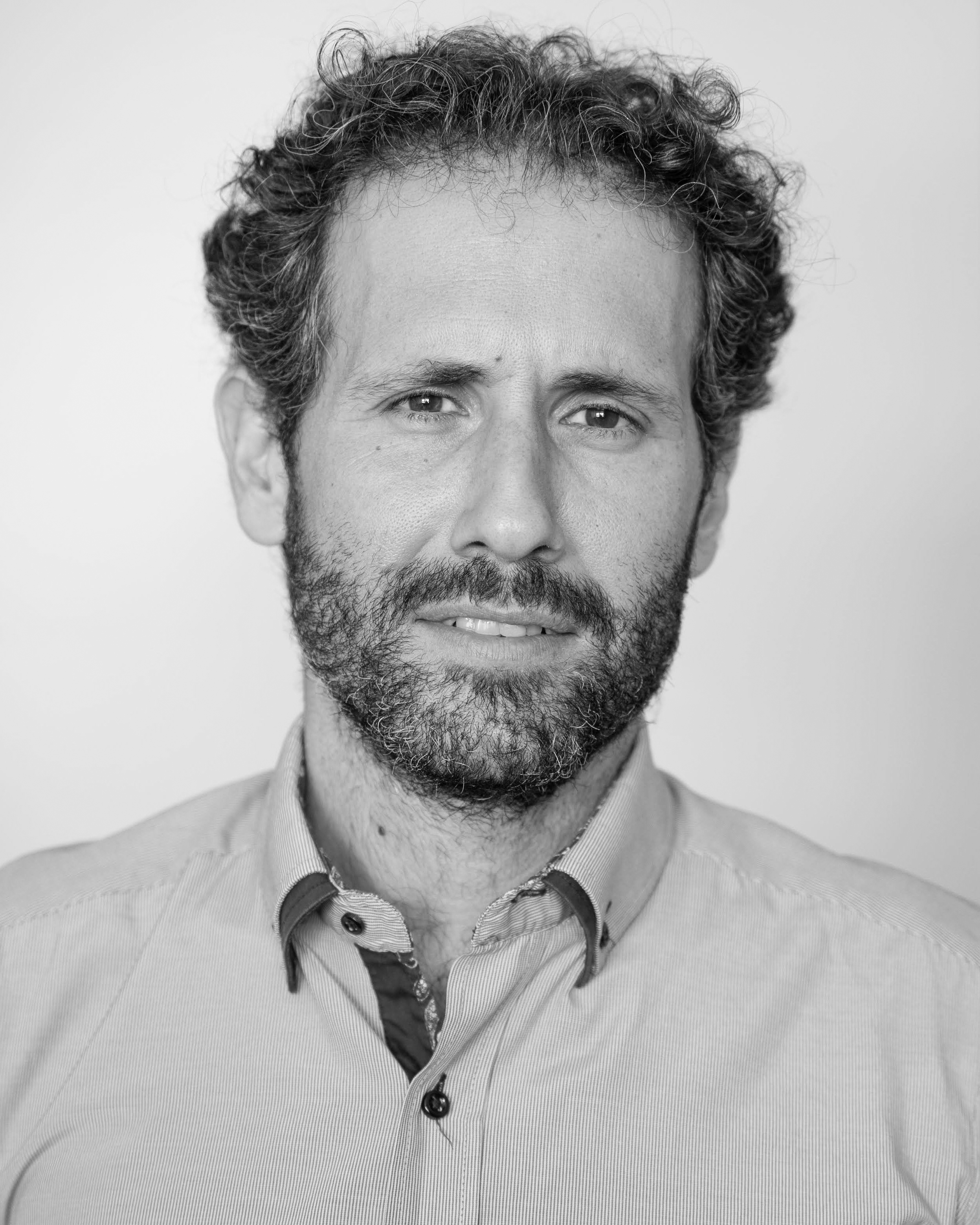}}]{Konstantinos Moustakas}
(Senior Member, IEEE) received the Diploma degree and the PhD in electrical and computer engineering from the Aristotle University of Thessaloniki, Greece, in 2003 and 2007 respectively. During 2007-2011 he served as a post-doctoral research fellow in the Information Technologies Institute, Centre for Research and Technology Hellas. He is currently a Professor at the Electrical and Computer Engineering Department of the University of Patras, Head of the Visualization and Virtual Reality Group, Director of the Wire Communications and Information Technology Laboratory and Director of the MSc Program on Biomedical Engineering of the University of Patras. He serves as an Academic Research Fellow for ISI/Athena research center. His main research interests include virtual, augmented and mixed reality, 3D geometry processing, haptics, virtual physiological human modeling, information visualization, physics-based simulations, computational geometry, computer vision, and stereoscopic image processing. He is a senior member of the IEEE, the IEEE Computer Society and member of Eurographics.
\end{IEEEbiography}

\begin{IEEEbiography}[{\includegraphics[width=1in,height=1.25in,clip,keepaspectratio]{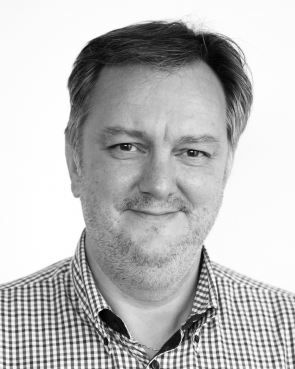}}]{Adrian Munteanu}
(Member, IEEE) is professor at the Electronics and Informatics (ETRO) department of the Vrije Universiteit Brussel (VUB), Belgium. He received the MSc degree in Electronics and Telecommunications from Politehnica University of Bucharest, Romania, in 1994, the MSc degree in Biomedical Engineering from University of Patras, Greece, in 1996, and the Doctorate degree in Applied Sciences (Summa Cum Laudae) from Vrije Universiteit Brussel, Belgium, in 2003. In the period 2004-2010 he was post-doctoral fellow with the Fund for Scientific Research – Flanders (FWO), Belgium, and since 2007, he is professor at VUB.
His research interests include image, video and 3D graphics compression, 3D video, deep-learning, distributed visual processing, error-resilient coding, and multimedia transmission over networks. 
%Use $\backslash${\tt{begin\{IEEEbiography\}}} and then for the 1st argument use $\backslash${\tt{includegraphics}} to declare and link the author photo.
%Use the author name as the 3rd argument followed by the biography text.
\end{IEEEbiography}



\vfill

\end{document}